\documentclass[journal]{IEEEtran}
\ifCLASSINFOpdf
\else
\fi
\usepackage{cite}
\usepackage{amsmath,amssymb,amsfonts}
\usepackage{algorithmic}
\usepackage{graphicx}
\usepackage{textcomp}
\usepackage{wrapfig}
\usepackage{float}
\usepackage{subfigure}

\usepackage{threeparttable}  
\usepackage{setspace}
\usepackage{multirow} 
\usepackage{multicol} 
\usepackage{color}

\usepackage{graphicx}

\usepackage{booktabs,makecell,multirow}
\usepackage{bm}

\usepackage{stfloats}



\begin{document}
%
\title{Cost-effective Mapping of Mobile Robot Based on the Fusion of UWB and Short-range 2D LiDAR}
%
%
%

\author{Ran Liu, Yongping He, Chau Yuen,~\IEEEmembership{Fellow,~IEEE,} Billy Pik Lik Lau, Rashid Ali, Wenpeng Fu, \\  and Zhiqiang Cao    
\thanks{This work is partially supported by the National Key R\&D Program of China 2019YFB1310805 and the Sichuan Science and Technology Program 2019YFH0161. (Corresponding author: Ran Liu).}

\thanks{Ran Liu,  Chau Yuen, and Billy Pik Lik Lau are with the Engineering Product Development Pillar, Singapore University of Technology and Design, Singapore 487372 (e-mail: ran\_liu@sutd.edu.sg; yuenchau@sutd.edu.sg; billy\_lau@mymail.sutd.edu.sg).} 

\thanks{Ran Liu, Yongping He,  Rashid Ali, Wenpeng Fu, and Zhiqiang Cao are with the School of Information Engineering, Southwest University of Science and Technology, Mianyang 621010, China (e-mail: yong\_ping\_he@163.com).}

\thanks{Rashid Ali is also with the Department of Computer Science, University of Turbat, Balochistan 92600, Pakistans.}
}

%
%

\markboth{IEEE/ASME Transactions on Mechatronics}%
{Shell \MakeLowercase{\textit{et al.}}: Cost-effective Mapping of Mobile Robot Based on the Fusion of UWB and Short-range 2D LiDAR}
%



\maketitle

\begin{abstract}
Environment mapping is an essential prerequisite for mobile robots to perform different tasks such as navigation and mission planning. With the availability of low-cost 2D LiDARs, there are increasing applications of such 2D LiDARs in industrial environments. However, environment mapping in an unknown and feature-less environment with such low-cost 2D LiDARs remains a challenge. The challenge mainly originates from the short-range of LiDARs and complexities in performing scan matching in these environments. In order to resolve these shortcomings, we propose to fuse the ultra-wideband (UWB) with 2D LiDARs to improve the mapping quality of a mobile robot. The optimization-based approach is utilized for the fusion of UWB ranging information and odometry to first optimize the trajectory. Then the LiDAR-based loop closures are incorporated to improve the accuracy of the trajectory estimation. Finally, the optimized trajectory is combined with the LiDAR scans to produce the occupancy map of the environment. The performance of the proposed approach is evaluated in an indoor feature-less environment with a size of $20m\times20m$. Obtained results show that the mapping error of the proposed scheme is 85.5\% less than that of the conventional GMapping algorithm with short-range LiDAR (for example Hokuyo URG-04LX in our experiment with a maximum range of 5.6$m$).
\end{abstract}


%
\IEEEpeerreviewmaketitle

\section{Introduction}
%
%
%
%
\IEEEPARstart{B}{UILDING} an environmental map is significant for autonomous robots to perform different tasks, including transportation, search, and rescue \cite{1,2}. The accuracy and quality of the map directly affect the ability of the robot to carry out its missions effectively. Reviewing the literature indicates that many researchers have proposed different solutions for mapping based on simultaneous localization and mapping (SLAM). In fact, SLAM has been a research hot spot in the field of robotics for decades. Efficient algorithms (including the approaches based on Kalman filter ~\cite{3}, particle filter~\cite{4}, and graph-based approach~\cite{5}) have been introduced for SLAM and the performance of these algorithms has been thoroughly evaluated in the literature.

In practical applications, different types of sensors such as inertial measurement unit (IMU), odometry, visual sensors, WiFi, LiDAR, and ultra-wideband (UWB) are used to implement a SLAM system. There is an economic incentive to use low-cost sensors with a moderate sensing capability to construct a precise map. IMU or odometry is an appropriate scheme to measure the position change over a short period. However, studies show that continuous use of them results in an unavoidable accumulative error~\cite{9}, which will lead to a serious deviation of the pose estimation and the tilt or distortion of the map. The visual-based approaches estimate the robot pose by matching features between the existing map and visual images~\cite{10}. However, visual SLAM is prone to errors, which originates from its sensitivity to light changes in the dark or low-textured environment~\cite{11}. The existing infrastructure deployed with WiFi network can be used for SLAM with low-cost hardware investment~\cite{12}. However, an analytical model is often required to describe the signal distribution in the environment~\cite{13}. 

Compared to the WiFi SLAM and visual SLAM, LiDARs are widely used in many SLAM systems due to their higher accuracy in measuring distance and superior performance under different light illuminations. GMapping \cite{15}, Hector SLAM \cite{16}, and Cartographer~\cite{17} are the most well-known and commonly used algorithms for 2D LiDAR-based SLAM. In GMapping algorithm, the Rao-Blackwellized particle filter (RBPF) is applied as an adaptive strategy for the grid mapping. Hector SLAM proposes a fast online construction of occupancy grid maps using a fast approximation of map gradients. One common disadvantage of Hector SLAM and GMapping algorithms is that their reliability for long-term mapping in feature-less environment is low. Cartographer is another widely adopted algorithm that fuses the information from LiDAR and IMU to eliminate the accumulated error by graph-based SLAM. However, the Cartographer algorithm is an expensive solution from the computational point of view. There are also many SLAM technologies based on 3D LiDAR, such as LiDAR Odometry and Mapping (LOAM) \cite{18}, LiDAR Inertial Odometry and Mapping (Lio-mapping) \cite{19}, and Surfel-based SLAM (Suma)~\cite{20}. However, the drawbacks of these methods are high computational complexity, and in this paper we focus on a low-cost solution.

LiDAR provides a precise pose estimation through scan matching in environments with rich features. However, it fails in environments with poor features. This shortcoming is especially more pronounced for the short-range low-cost 2D LiDARs (e.g., a maximum range of 5.6$m$ in an environment with a size of $20m\times20m$). Moreover, it is an enormous challenge for the robot to identify the same place in feature-less environment with LiDARs, and the accuracy and quality of the map is seriously reduced with the use of low-cost 2D LiDAR due to the difficulty of loop closure detection. UWB technology provides a low-cost way to measure the distance with high accuracy (i.e., the ranging can be up to 20m with error less than 0.1m)~\cite{21}. Due to the unique ID and high ranging accuracy, UWB offers a solution for accurate positioning in feature-less environment through wireless technology.

Therefore, in this article, we present a novel approach, which incorporates UWB ranging, odometry, and LiDAR measurements to estimate the trajectory of a mobile robot and construct the precise occupancy map in an unknown environment. In our approach, UWB provides an approximate estimate of the position without the need to perform loop closure detection. Do note that in our solution, there is no need to know the location of the deployed UWB nodes, they just have to be remained static throughout the mapping period. It should be indicated that unlike LiDAR, each UWB node can provide a unique ID for the identification. It is expected that the combination of UWB and LiDAR features can lead to a robust solution to generate the map in an unknown feature-less environment. Main contributions and advantages of the present article can be summarized as follows:

\begin{itemize}
	
	\item A novel solution is proposed to incorporate odometry, UWB, and LiDAR information into a two-step graph optimization framework for performing a robust mapping in an unknown and feature-less environment. 
	
	\item An approach to perform scan matching based on local-submaps, which consist of multiple continuous LiDAR scans, is presented to improve the accuracy of short-range 2D-LiDAR loop closure detection.
	
	\item Performance of the proposed approach is thoroughly evaluated in one building at the campus with an area of about 400m$^2$. Obtained results show that compared to the conventional GMapping algorithm, the proposed system can generate a better map at a low cost. 
\end{itemize}

Contents of the present article are organized as follows: A comprehensive literature survey is presented in Section~\ref{Related Work}. Afterwards, the proposed system and details of the method are illustrated in Section~\ref{Mapping system}. Results and discussions are presented in Section~\ref{Experimental Results} and finally, main conclusions and future works are summarized in Section~\ref{Conclusion}.

\section{Related Work}
\label{Related Work}
The robot should be capable of identifying free spaces and obstacles to navigate precisely and safely with a consistent map. To this end, the robot should be capable of self-exploring and constructing the map of an unknown environment. Recent literatures show a growing interest in simultaneous localization and mapping (SLAM) using mobile robots in unknown environments. It is worth noting that there are various implementations with different sensors such as the LiDAR and depth cameras~\cite{22}. Depending on the sensor type, sensors can be divided into two most widely used SLAM techniques, namely visual SLAM and LiDAR SLAM techniques~\cite{23,24,25}.

Visual SLAM is implemented by utilizing visual features as landmarks for pose estimation. Parallel tracking and mapping (PTAM) introduces to split camera tracking and mapping with two threads~\cite{26}. RGB-D cameras provide 3D information in real time, but they are sensitive to the ambient lighting. Studies showed that such shortcomings make it tough for vision-based modeling of the environment~\cite{28}. Meanwhile, the analysis of images has high computational complexity. Compared with visual sensors, LiDAR provides measurement with superior characteristics such as high reliability, better precision, and robust to variations in illumination conditions. Therefore, LiDAR-based SLAM is widely adopted as the most stable and reliable SLAM solution~\cite{29}. LiDAR achieves a low-drift motion with reasonable computational complexity. However, the LiDAR odometry heavily relies on scan matching, which requires a long range LiDAR to scan the surrounding and shows poor performance in feature-less environments~\cite{30}. 

Though the positioning accuracy of GPS degrades due to the occlusion of signals by obstacles (for example, tunnels and buildings), it provides a very good positioning accuracy in outdoor environment without the block of signals. This motivates one to use GPS to support the SLAM when the satellite signals are available. GPS-guided SLAM ~\cite{31} uses the position estimated from GPS to correct the dead reckoning error, while our approach treats the UWB as landmarks for the correction of dead reckoning and the location of UWB nodes does not need to be known in advance.

Considering the limitations of each type of sensors, scholars have designed diverse mapping approaches using multi-sensor data. In the multi-sensor information fusion technology, collected data from different sensors are processed to reduce uncertainty and achieve a consistent description of the environment~\cite{32}. Liu et al.~\cite{35} presented a system that fuses the pedestrian dead reckoning and received signal strength (RSS) measurements from the surrounding WiFi access point to estimate the trajectory of multiple users. Wen et al.~\cite{36} presented a backpack mobile mapping system that is mainly designed for indoor applications. Hao et al.~\cite{37} provided an efficient and stable system to detect glass in the environment based on the fusion of ultrasonic and LiDAR. Maddern et al.~\cite{38} proposed a probabilistic model for fusing sparse 3D LiDAR information with stereo images to obtain reliable depth maps in real-time estimates. However, further investigations showed that the proposed model requires complex recognition algorithms so that a pre-training stage and high computational expense are unavoidable. Luo et al.~\cite{39} combined a stereo camera, laser range finder, and wheel odometry to construct an enriched indoor map. 

The hardware investments and the accuracy requirements of indoor maps are very diverse, which mainly originates from different budgets and applications. Accordingly, researchers fused the UWB sensor with other low-cost sensors to obtain an accurate robot pose and improve the accuracy and quality of mapping economically. Wang et al.~\cite{40} combined UWB and visual-inertial odometry to overcome the visual drift through UWB range measurements and improve robustness of the system. Song et al.~\cite{41} proposed a 2D range-only SLAM method by incorporating low-cost LiDAR and UWB sensors to dynamically localize the robot and beacons, and create a 2D map of an unknown environment in real time. Their experiments demonstrated that the fusion of UWB and LiDAR enables drift free SLAM in real-time based on ranging measurements only. In order to realize robust mapping in our study, UWB ranging information and odometry are fused into a graph optimization framework to obtain the initial trajectory. Then the LiDAR loop closure information is applied to further optimize trajectory of the robot.
\begin{figure}[H]
	\centering
	\includegraphics[width=9cm]{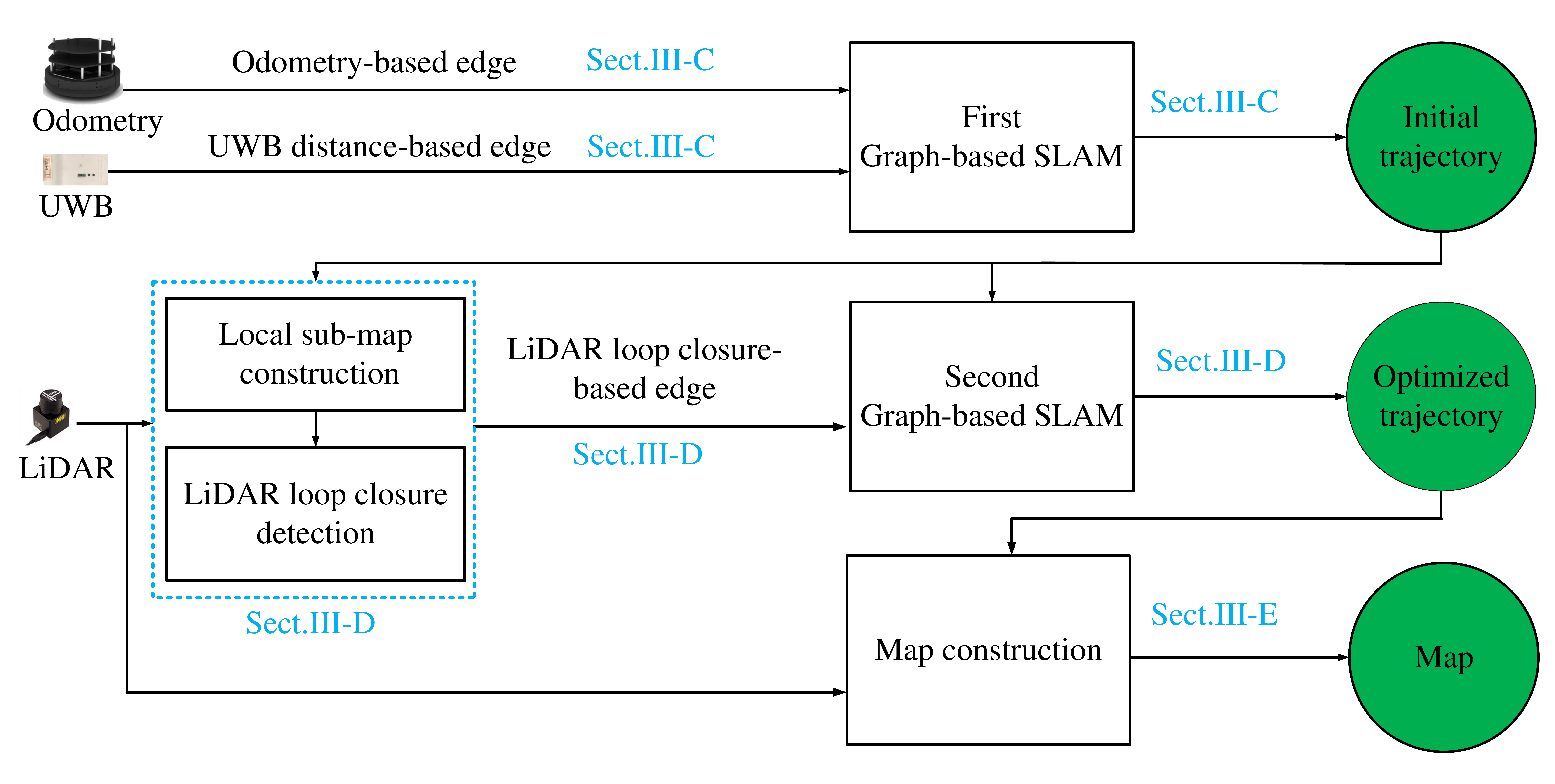}
	\caption{System overview of the proposed approach. }
	\label{figure:system_overview1}	
\end{figure}
\section{Mapping Based on Fusion of UWB and LiDAR}
\label{Mapping system}

\subsection{System Overview}
The mobile robot is equipped with different sensors to perceive the environment. Wheel odometry is used for self localization based on a kinematics model and encoder. LiDAR is used to measure the distance to the surrounding objects. Moreover, UWB sensor is carried by the robot to obtain the ranging information from the UWB nodes in proximity. This article mainly focuses on constructing a map of a moving robot carrying a UWB node, LiDAR, and odometry.

Fig.~1 shows that UWB ranging information, odometry, and LiDAR information are fused into a graph optimization framework. More specifically, the UWB ranging information is initially fused with odometry to obtain the initial trajectory of the robot. The pose from the odometry and UWB nodes are considered as vertices in the graph. UWB distance-based and odometry-based edges are used as constraints during the first optimization. The former is represented by the distance from the robot to the UWB node, while the latter is represented by the pose transformation between adjacent vertices. The UWB information can correct the accumulated error caused by the odometry to produce a coarse optimized robot trajectory during the first graph optimization. It is important to point out that the location of the UWB node does not need to be known. In order to improve the quality of the map, we further incorporate the LiDAR-based edges in a second pose graph optimization. These edges are determined by scan matching based on local sub-map matching using iterative closest point (ICP). Through the fusion of three types of edges, a more accurate robot trajectory can be obtained. Finally, the optimized robot trajectory is combined with the LiDAR data to generate the occupancy map of the environment.

A 2D map of the environment is created and the position of obstacles in z-axis is not considered, as only a 2D LiDAR is used for the experiment. We are not able to construct the 3D map of the environment, since the height information of obstacles is not available through the 2D LiDAR. Though we use only low-cost 2D LiDAR to demonstrate the idea, the same idea can also be applied for 3D LiDAR. Our approach can be extended to a 3D LiDAR for 3D pose estimation and creating a 3D map of the environment.

\subsection{Graph-based SLAM in General}

In the Graph-based SLAM, vertices are represented by robot poses. In addition, we consider the UWB node as vertex (with unknown location) in the graph. Edges consist of relationships between these vertices. The main purpose of the graph optimization is to adjust vertices to better satisfy the edges. Our goal is to optimize the robot pose based on the fusion of odometry-based, UWB-based, and LiDAR-based edges. Let $X$=($X_1, ... ,X_T, U_1, ..., U_M$) be a vector of parameters, where $X_i$ and $X_j$ denote the pose of the robot (including the 2D location and heading) at time $i$ and time $j$, and $U_m=(x_{UWB_m},y_{UWB_m})$ denotes the 2D location of the UWB node. Moreover, $m$ is the unique ID of UWB nodes. \iffalse Furthermore, it is assumed that $Z_{i,j}$, $Z_{i^{m}}$ and $\Omega_{i}^j$, and $\Omega_{i}^m}$ represent the mean and the information matrix of measurement between vertex $i$, vertex $j$ and vertex $m$ , respectively.\fi Constraints are additionally parameterized with a certain degree of uncertainty, which is denoted as the information matrix (i.e., $\Omega_{{i-1}}^i$, $\Omega_{i}^j$ and $\Omega_{i}^m$). $C$ represents  the set of LiDAR loop closure constraints. It should be indicated that the main purpose of the maximum likelihood approach is to find the configuration of vertices $X^{*}$ to meet the following criteria:

{
\footnotesize
\begin{equation}
\begin{array}{l}
\underset{X}{\arg \min } \underbrace{\sum_{(i, j) \in C}^{t}\left(Z_{i}^{j}-\tilde{Z}_{i}^{j}\left(X_{i}, X_{j}\right)\right)^{T}\Omega_{i }^j\left(Z_{i}^{j}-\tilde{Z}_{i}^{j}\left(X_{i}, X_{j}\right)\right)}_{\text {LiDAR loop closure-based constraint }}+ \\
\underbrace{\sum_{i=2}^{t}\left(Z_{i-1}^{i}-\tilde{Z}_{i-1}^{i}\left(X_{i-1}, X_{i}\right)\right)^{T}\Omega_{i-1}^ i\left(Z_{i-1}^{i}-\tilde{Z}_{i-1}^{i}\left(X_{i-1}, X_{i}\right)\right)}_{\text {Odometry-based constraint }} \\
+\underbrace{\sum_{i=1}^{t} \sum_{m=1}^{M}\left(Z_{i}^{m}-\tilde{Z}_{i}^{m}\left(X_{i}, U_{m}\right)\right)^{T}\Omega_{i}^m\left(Z_{i}^{m}-\tilde{Z}_{i}^{m}\left(X_{i}, U_{m}\right)\right)}_{\text {UWB distance-based constraint }}
\end{array}
\footnotesize
\end{equation}
}where $Z_{i-1}^{i}$, $Z_{i}^j$, and $Z_{i}^m$ are the actual observations between two vertices, including the pose transformation between adjacent vertices $X_{i-1}$ and $X_i$ represented by the odometry, the pose transformation between non-adjacent robot vertices $X_i$ and $X_j$ represented by the LiDAR loop closure, and the distance between robot pose $X_i$ and UWB node $UWB_m$. Furthermore, $\tilde{Z}_{i-1}^{i}(X_{i-1},X_i)$ is the predicted odometry between the adjacent vertices $X_{i-1}$ and $X_i$. $\tilde{Z}_{i}^j(X_i,X_j)$ is the prediction of pose transformation based on the current configuration of non-adjacent nodes $X_i$ and $X_j$. $\tilde{Z}_{i}^m(X_i,U_m)$ denotes the prediction of a distance between robot pose $X_i$ and UWB node $U_m$. Fig. 2 illustrates the vertices and edges in the graph structure based on the proposed approach.

\begin{figure}[H]
	\centering
	\includegraphics[width=6.5cm]{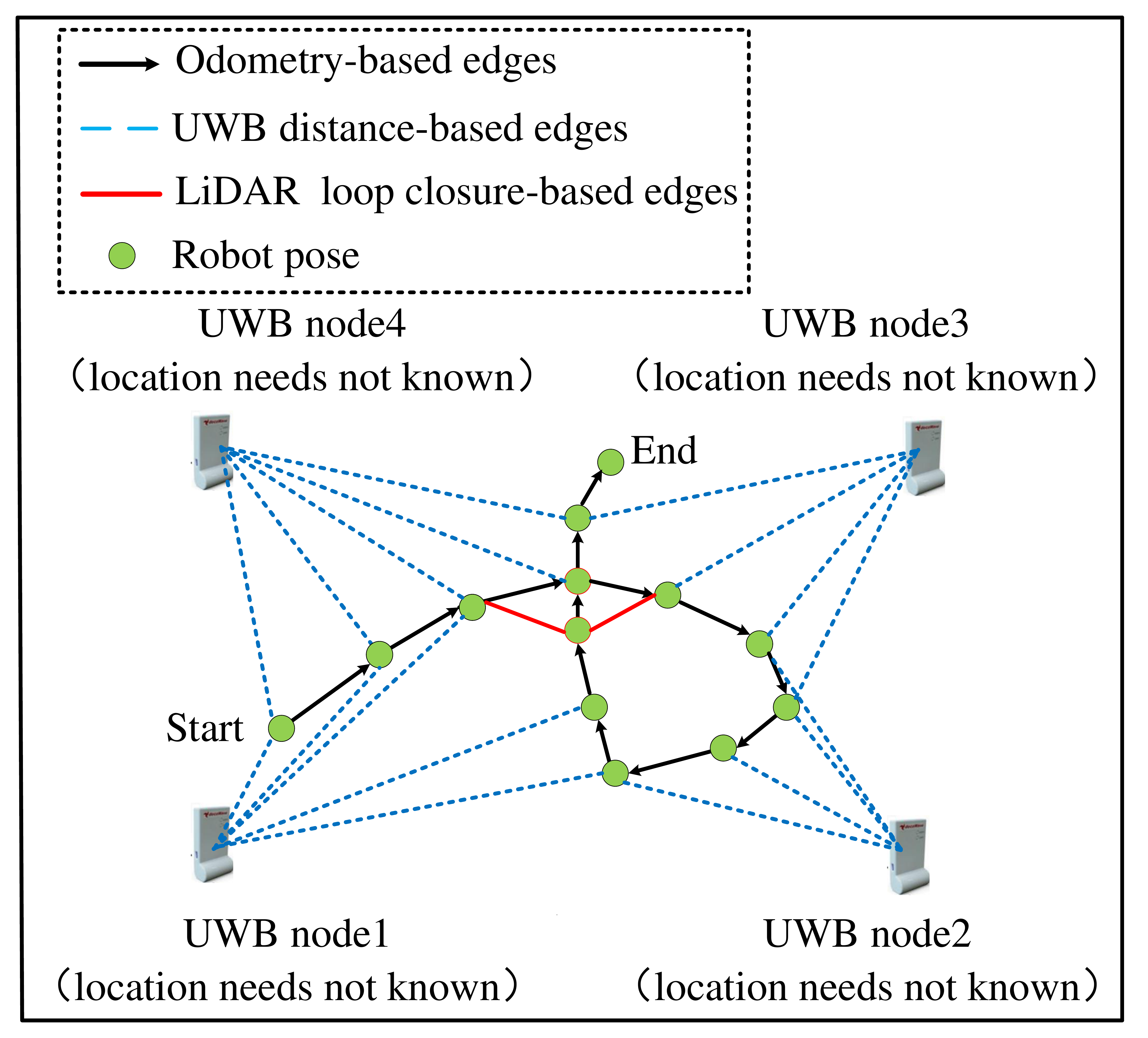}
	\caption{Vertices and edges in the graph.}
	\label{figure:graph-structure}
\end{figure}

\subsection{First Optimization Based on UWB and Odometry}  

In this section, the UWB ranging measurement is fused with odometry to correct the robot pose during the first optimization. Therefore, it is essential to construct the UWB distance-based and odometry-based edges during this stage.
\subsubsection{Odometry-based Edge} 

In particular, for graph-based SLAM, $Z_{i}^j$ is known as edge, which represents a rigid-body transformation between vertices $X_i$ and $X_j$. The odometry-based edge is determined based on the relative translation and rotation between the sequential odometry measurements. The rigid-body transformation between the pose at time $i-1$ and time $i$ can be calculated by:
\begin{equation} \small
\left[\begin{array}{ccc}
\triangle x\\
\triangle y\\
\triangle \theta
\end{array}
\right]
=\left[
\begin{array}{ccc}
cos(\theta_{i-1}) & -sin(\theta_{i-1}) & 0 \\
sin(\theta_{i-1}) &  cos(\theta_{i-1}) & 0\\
0 &0 &1
\end{array}
\right] \left[
\begin{array}{ccc}
x_i-x_{i-1}  \\
y_i-y_{i-1}  \\
\theta_i-\theta_{i-1}
\end{array}
\right] ,
\end{equation}

\subsubsection{UWB Distance-based Edge}
UWB ranging is applied to construct the UWB distance-based edge for removing cumulative errors of the odometry. The distance measured from UWB constitutes UWB distance-based edge. For example, when we obtain a UWB ranging measurement $d_{ij}$ from UWB node $j$ at pose $X_i$, an edge $<i, j>$ is added to the graph.

\subsection{Second Optimization by Additional LiDAR Constraints}
The LiDAR loop closure-based edge between non-adjacent vertices is added to further constrain the pose during the second pose graph optimization after the first optimization. Studies show that loop closure detection is essential for graph optimization method~\cite{42}. In other words, the correct LiDAR loop closure detection can remove the cumulative error of the odometry, thereby obtaining a consistent map. Conversely, wrong loop closures interfere with the subsequent map optimization and may even destroy the existing map creation.

\subsubsection{Local Sub-map Construction}
When the robot re-enters a known area after moving for a long time, the loop closure detection algorithm searches the stored database to match the current LiDAR scan. Based on the similarity between two scans, which is obtained through the ICP algorithm, it is possible to determine whether the robot has visited the same scene. Studies show that the conventional ICP algorithm has several drawbacks for detecting loop closure with short-range LiDARs in feature-less environment. In order to resolve these shortcomings, local sub-maps are constructed for detecting the potential loop closures. Local sub-maps contain multiple frames of LiDAR scans, thereby eliminating the limitations of the conventional ICP algorithm, which uses the single LiDAR scan to determine the LiDAR loop closure. The local map is constructed when the moving distance of the robot is less than a certain threshold of $\epsilon$. Detection of the LiDAR loop closure is achieved by matching the local maps through ICP.

Each local sub-map involves a large number of LiDAR scan points that should be processed accordingly to shorten the required time for the mapping process. For raw LiDAR data, the voxel filtering is initially applied to reduce the number of points. Then the statistical filtering is used to perform statistical analysis on the neighborhood of each point and remove LiDAR points that do not meet the standard. Finally, the radius filtering is utilized to further reduce the number of noisy points. Fig.~\ref{figure:local map} shows examples of the LiDAR local sub-maps and LiDAR single scans based on initial trajectory after applying UWB-based edges.

\begin{figure*}[htbp]
	\centering
	\subfigure[Example of the local sub-map based on the initial trajectory]{
		\label{figure:local map}
		\includegraphics[width=0.49\linewidth]{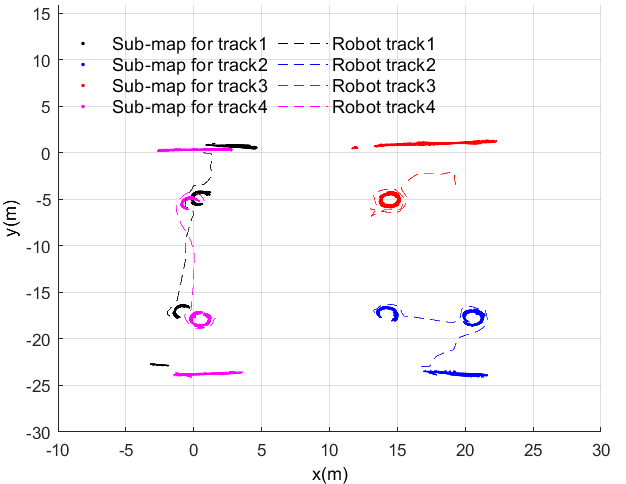}
	}
	\subfigure[Example of the single LiDAR scan based on the initial trajectory]{
		\label{figure:singlescan}
		\includegraphics[width=0.48\linewidth]{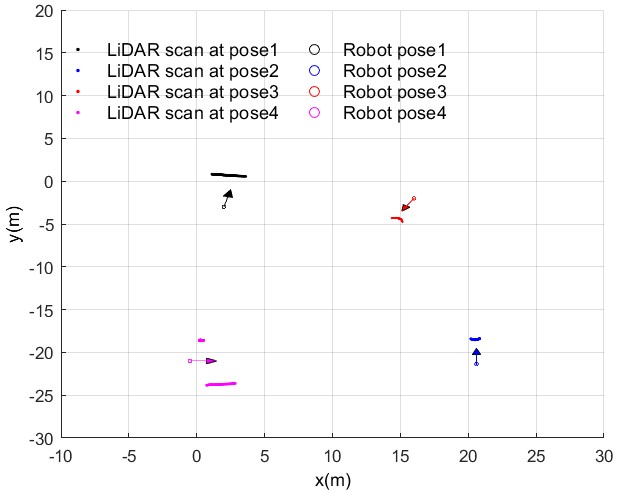}}
	\caption[LiDAR loop]{Comparison of the local sub-map and single LiDAR scan.
	}
\vspace{-0.2cm}
	\label{figure:local map}
\end{figure*} 

\subsubsection{LiDAR Loop Closures Detection Based on Local Sub-map Matching}
\begin{figure*}[htbp]
	\centering
	\subfigure[Example of the raw track (light blue color), the initial trajectory (black color), and the LiDAR-based loop closures (pink color)]{
		\label{figure:LOOP1}
		\includegraphics[width=0.511\linewidth]{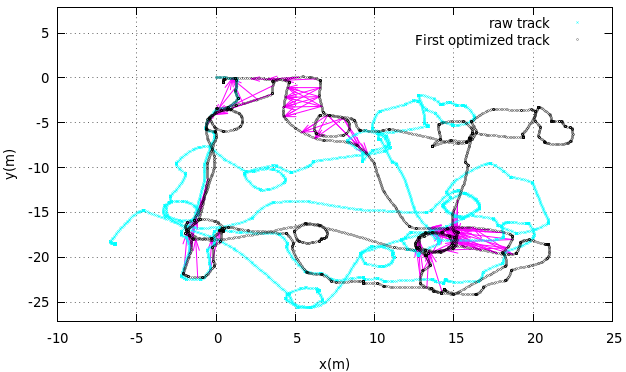}
	}
	\subfigure[The registered result (blue color) between source LiDAR (red color) and target LiDAR (green color)]{
		\label{figure:LOOP2}
		\includegraphics[width=0.45\linewidth]{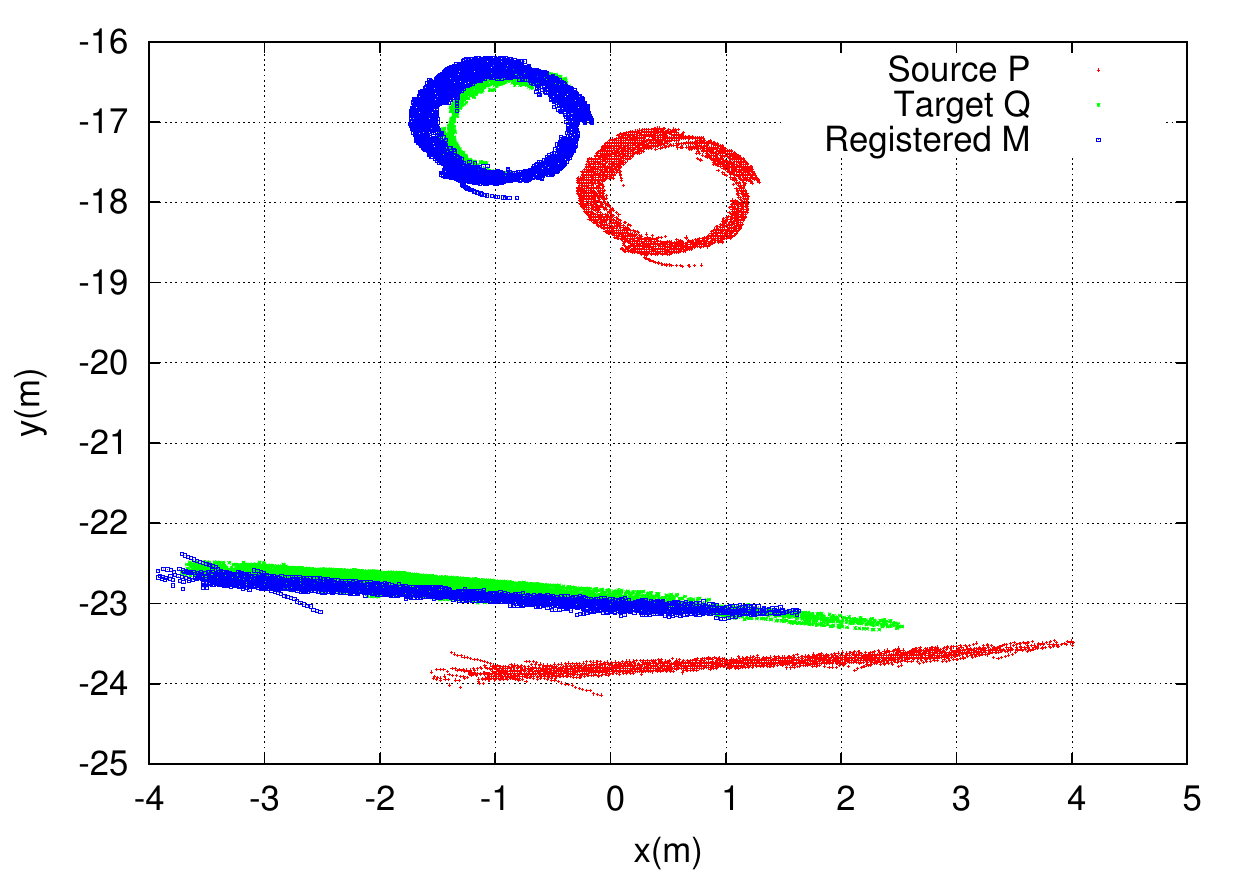}}
	\caption[LiDAR loop]{ LiDAR loop closure detections based on the local sub-map matching.
	}
	\label{figure:The LiDAR loop closure}
	\vspace{-0.2cm}
\end{figure*}

The ICP algorithm is an efficient least-square registration process, which can achieve optimal rigid body transformation by selecting appropriate point pairs and conducting iterative calculations until the convergence of the numerical solution. Assuming we have two local LiDAR point cloud sets, i.e., the source point cloud $Q=(q_1...q_{N})$ and the target point cloud $P=(p_1...p_{N})$, the nearest neighbor method is used to match the points in both point cloud sets. The goal is to rotate and translate the point cloud $P$ to minimize the objective function $E(R,T)$:
\begin{equation} 
\label{1}
E(R,T)=\dfrac{1}{N}\sum_{i=1}^N  \left\|q_i-R \cdot p_i-T\right\|^{2},
\end{equation}
where $N$ is the number of the corresponding LiDAR points in the local sub-maps. Moreover, $R$ and $T$ denote the rotational and translational matrices, respectively. It should be indicated that $R$ and $T$ are substituted into the source LiDAR point cloud to obtain a new point set $M$. Then, $M$ is matched with the target point cloud $Q$ to solve the new rotational and translational matrices. This numerical solution converges after several iterations. If the distance between the corresponding LiDAR points reaches a threshold value $d$, it is identified as outlier and discarded by the algorithm. The iterative calculation ends when the solution converges. It should be indicated that the fitness score $E(R,T)$ determines the average Euclidean distance between the corresponding LiDAR points after registration. If the fitness score is smaller than a predefined threshold $\sigma$, $<X_i, X_j>$ is considered as the LiDAR-based loop. Fig.~\ref{figure:LOOP1} shows the LiDAR loop closures during the second optimization. Moreover, Fig.~\ref{figure:LOOP2} illustrates the registered result of the source and target LiDAR sub-maps.

\subsection{Pose Graph Optimization and Map Construction}
Equation 1, which represents a graph consisting of poses and edges, is finally optimized through the pose graph optimization algorithm. The Levenberg-Marquardt solver in g2o is used to optimize the graph~\cite{43}. We use the optimized trajectory and LiDAR scans to construct the grid map of the environment, where each grid is a binary random variable that specifies the occupancy. Moreover, the Bresenham algorithm~\cite{44} is used to calculate the set of non-obstacle grid points based on the robot pose and the LiDAR scans in the grid map. It is worth noting that occupancy probability of each grid is updated through the binary Bayesian filter~\cite{45}.

\section{Experimental Results}
\label{Experimental Results}
\subsection{Experimental Design}

\begin{figure*}[htbp]
	\centering
	\subfigure[Overall layout of the experiment environment]{
		\label{figure:overall}
		\includegraphics[width=0.7\linewidth]{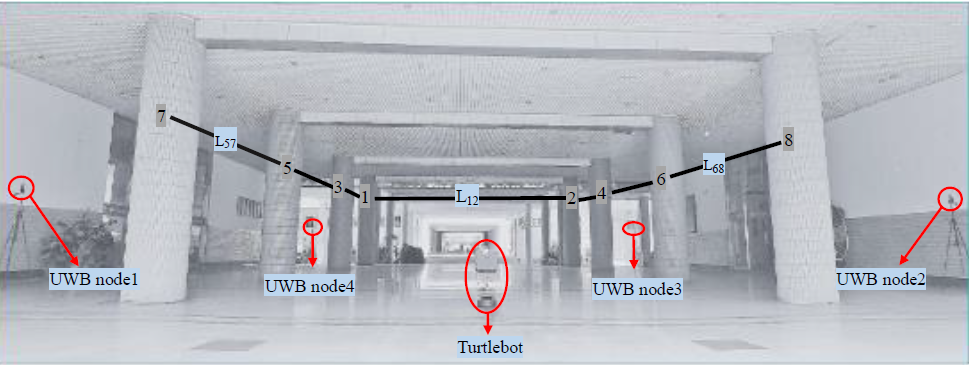}
	}
	\subfigure[Sensors equipped by the robot]{
		\label{figure:robot platform}
		\includegraphics[width=0.27\linewidth]{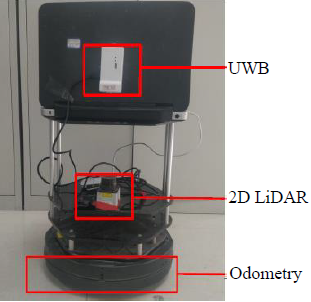}}
	\caption[LiDAR loop]{Experimental setup.
	}
	\label{figure:experient}
\end{figure*}

In order to evaluate the performance of the proposed approach, experiments are conducted in an indoor building with a size of 400m$^2$. Fig.~\ref{figure:overall} shows the overall layout of the experiment. A mobile robot TurtleBot-II is equipped with UWB sensors, wheel odometry, and Hokuyo laser ranger finder to explore the environment. Fig.~\ref{figure:robot platform} shows the sensors installed on the robot. In our approach, the environment is completely unknown. The robot is manually controlled to collect UWB ranging and LiDAR scans. The goal is to simultaneously localize the robot and map the environment (including the LiDAR map of the environment and the locations of the UWB nodes) when exploring the environment.

\def\degree{${}^{\circ}$}

In the experiment, DecaWave DWM1001 with a maximum range of 20m and an average ranging accuracy of 0.1m is used for UWB distance ranging. Moreover, Hokuyo URG-04LX (maximum effective range of 5.6m) is utilized for LiDAR ranging measurements. The LiDAR has a scanning angle of 240\degree\ with an angular resolution of 0.36\degree. The price of our short-range 2D LiDAR is about 930 USD and each UWB node costs about 30 USD. The total cost of five UWB nodes (one node on the robot and four nodes are placed as landmarks in the environment) and a Hokuyo URG-04-LX LiDAR for our experiment is about 1080 USD. However, a long-range 2D LiDAR (for example Hokuyo UST-20LX with a maximum range of 20m) costs about 2600 USD. Thus, our configuration reduces the cost when compared to the traditional solutions using the long-range 2D LiDARs. In the present study, UWB nodes are randomly placed in the test environment, and the locations of UWB nodes need not known. During the experiment, the robot moves with an average speed of 0.2m/s for 31 minutes in the environment. 

In Fig. 5, we marked the eight columns in the environment.  Four columns, including those marked with 1, 3, 5, and 7 from top to bottom, are on the left side, while the other four columns marked by 2, 4, 6, and 8, are on the right side of the environment. $L_{ab}, ({a,b}\in(1-8))$ denotes the distance between the center of the column $a$ and $b$. Distance between two columns is measured through the RVIZ tool in robot operating system (ROS). In the experiment, mapping error between the estimated and real distance is initially calculated. Then, mean error is calculated as a representation of the mean mapping error to evaluate the measuring accuracy. The UWB used in our experiment has an average ranging accuracy of 0.1m under ideal conditions with clear line of sight. It is difficult to achieve a ranging accuracy of 0.1m in practical environment with obstacles, due to the non-line-of-sight effect. As a result, a positioning accuracy of 0.1m is not guaranteed when UWB is used for localization.

\renewcommand{\arraystretch}{1.326} 
\begin{table*}[ht] \huge 
	\centering 
	\caption{Mapping error for different algorithms.}
	\label{tab:Mapping error}  
	\scalebox{0.779}{	
		\scalebox{0.48}{
			\begin{tabular}{|c|c|c|c|c|c|}
				
				\hline
				\multirow{6}{*}{\textbf{{\Huge Mapping error}}} & \multicolumn{5}{c|}{\textbf{{\Huge Method}}}                                                                                                                                                                     \\ \cline{2-6} 
				& \multicolumn{1}{l|}{\textbf{\begin{tabular}[c]{@{}c@{}}{\Huge Gmapping} \\ ({\Huge odometry}  \\  {\Huge and LiDAR})\end{tabular}}} & \multicolumn{1}{l|}{\textbf{\begin{tabular}[c]{@{}c@{}}{\Huge Odometry and LiDAR} \\ {\Huge (no LiDAR} \\ {\Huge loop closure)}\end{tabular}}} &
				\multicolumn{1}{l|}{\textbf{\begin{tabular}[c]{@{}c@{}}{\Huge Odometry, LiDAR},\\ {\Huge and  LiDAR} \\{\Huge loop closure}\end{tabular}}}
				& \multicolumn{1}{l|}{\textbf{\begin{tabular}[c]{@{}c@{}}{\Huge (Proposed)} \\ {\Huge Odometry and UWB}  \\  {\Huge with only}\\ {\Huge first optimization}\end{tabular}}}
				& \multicolumn{1}{l|}{\textbf{\begin{tabular}[c]{@{}c@{}}{\Huge (Proposed)}\\{\Huge Odomery, UWB,}\\ {\Huge and LiDAR} \\  {\Huge with first and} \\ {\Huge second optimization}\end{tabular}}} \\ [10pt] \hline 
				{\Huge $L_{12}$}                    & {\Huge2.802}                 & {\Huge5.391}                         & {\Huge1.447}                                    & {\Huge0.465}                                 & {\Huge0.535}                                        \\ 
				{\Huge $L_{34}$}                   & {\Huge2.489}                         & {\Huge2.961}                         & {\Huge0.211}                                    & {\Huge0.216}                                 & {\Huge0.096}                                        \\ 
				{\Huge $L_{56}$}                   & {\Huge1.000}                         & {\Huge0.785}                        & {\Huge0.785}                                    & {\Huge0.078}                                & {\Huge0.267}                                        \\ 
				{\Huge$L_{78}$}                    & {\Huge0.573}                         & {\Huge1.092}                         & {\Huge2.411}                                    & {\Huge0.086}                                & {\Huge0.023}                                        \\ 
				{\Huge$L_{13}$}                    & {\Huge0.160}                         & {\Huge0.184}                         & {\Huge0.065}                                    & {\Huge0.381}                                 & {\Huge0.073}                                        \\ 
				{\Huge$L_{35}$}                   & {\Huge1.196}                         & {\Huge1.113}                         & {\Huge0.004}                                    & {\Huge0.100}                                 & {\Huge0.004}                                        \\ 
				{\Huge$L_{57}$}                    & {\Huge0.880}                         & {\Huge1.644}                         & {\Huge0.234}                                    & {\Huge0.888}                                 & {\Huge0.187}                                        \\ 
				{\Huge$L_{24}$}                    & {\Huge1.421}                         &{\Huge0.051}                         & {\Huge0.120}                                    & {\Huge0.083}                                 & {\Huge0.025}                                       \\ 
				{\Huge$L_{46}$}                   & {\Huge0.979}                         &{\Huge0.179}                         & {\Huge0.138}                                    & {\Huge0.616}                                 & {\Huge0.283}                                        \\ 
				{\Huge $L_{68}$}                  & {\Huge0.750}                        & {\Huge0.304}                        & {\Huge0.112}                                    & {\Huge0.112}                                & {\Huge0.499}                                      \\ 
				{\Huge Mean mapping error(m)}          & {\Huge1.225}                         & {\Huge1.372}                         & {\Huge0.623}                                    & {\Huge0.301}                                 & {\Huge0.199}                                      \\ \hline
			\end{tabular}
	}}
\end{table*}

\subsection{Map Comparison with Different Algorithms}

\begin{figure}[]
	\centering
	\quad
	\subfigure[GMapping (odometry and LiDAR)]{
		\label{figure:GMapping}
		\includegraphics[width=0.27\linewidth]{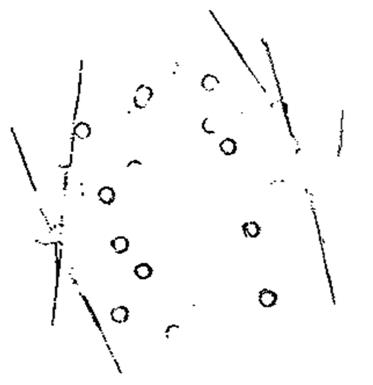}
	}
	\quad
	\subfigure[Odometry and LiDAR (no LiDAR loop closure)]{
		\label{figure:Odom-only}
		\includegraphics[width=0.27\linewidth]{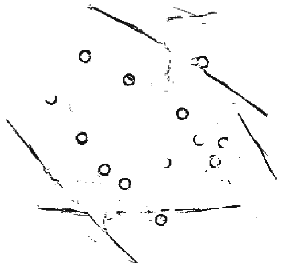}
	}
	\centering
	\quad
	\subfigure[Odometry, LiDAR, and LiDAR loop closure]{
		\label{figure:Odom-laserloop}
		\includegraphics[width=0.25\linewidth]{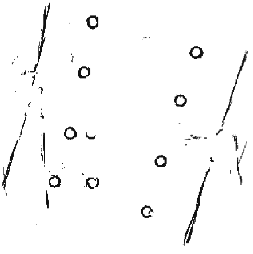}
	}
	\quad
	\subfigure[(Proposed) Odometry and UWB with only first optimization]{
		\label{figure:Odom-uwb}
		\includegraphics[width=0.25\linewidth]{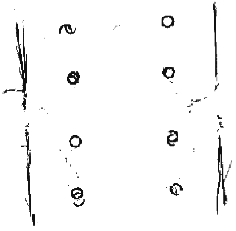}
	}
	\quad
	\subfigure[(Proposed) Odomery, UWB, and LiDAR with first and second optimization]{
		\label{figure:Odom-uwb-laserloop}
		\includegraphics[width=0.25\linewidth]{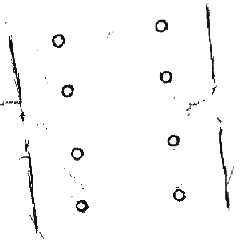}
	}
	\quad
	\caption[mapping results]{Obtained maps from different algorithms.
	}
	\label{figure:Mapping result}
\end{figure} 

Fig.~\ref{figure:Mapping result} and Table~\ref{tab:Mapping error} show the comparison of the map quality and accuracy for different algorithms. It indicates that the average mapping error for the GMapping algorithm (i.e., fuses the odometry and LiDAR information) is 1.225m. It is found that the maps presented in Fig.~\ref{figure:GMapping} and Fig.~\ref{figure:Odom-only} are quite different from the real experiment environment. This may be attributed to large cumulative error of the odometry as the robot moves on uneven ground for a long time. The inaccurate estimation of the robot pose leads to the drift of the wall and eight columns in the corresponding maps. Fig.~\ref{figure:Odom-laserloop} shows the map based on the fusion of odometry and LiDAR loop closures. In this case, the wall still has a large deviation from the actual position. This is because the cumulative error of the odometry is not corrected without the integration of UWB. However, there are six columns in Fig.~\ref{figure:Odom-laserloop} obtained from the LiDAR-based loop closure, which is consistent with the real columns in the environment. Fig.~\ref{figure:Odom-uwb} shows the created map by combining UWB and odometry. It is observed that the map is roughly in line with the real map. The mapping error with the fusion of UWB and odometry is reduced by 78.1\% (i.e., from 1.372m to 0.301m) when it is compared with the case with odometry alone. It is a challenge for a robot to identify the same place in feature-less environment with 2D low-cost LiDAR. Therefore, eight columns in Fig.~\ref{figure:Odom-uwb} are poorly mapped because LiDAR loop closures are not used to constrain the robot trajectory. Fig.~\ref{figure:Odom-uwb-laserloop} shows the best mapping result with a mapping error of 0.199m after fusing odometry, UWB, and LiDAR. In this case, the mapping accuracy is improved by 85.5\% and 33.9\% when compared to the case of odometry alone (with a mapping error of 1.372m, see Fig.~\ref{figure:Odom-only}) and the case without LiDAR loop closures (with a mapping error of 0.301m, see Fig.~\ref{figure:Odom-uwb}), respectively. It is found that when LiDAR and UWB are fused, UWB sensors can be considered as additional attributes, which provide precise pose estimation of the robot. On the other hand, LiDAR provides a good representation of the environment and helps to improve the pose estimation obtained from UWB.

\begin{figure}[]
	\centering
	\subfigure[UWB1]{
		\label{figure:UWB1}
		\includegraphics[width=0.29\linewidth]{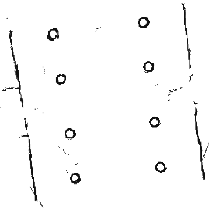}
	}
	\subfigure[UWB2]{
		\label{figure:UWB2}
		\includegraphics[width=0.29\linewidth]{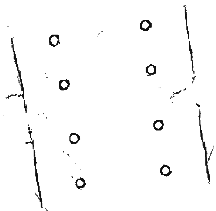}
	}
	\subfigure[UWB3]{
		\label{figure:UWB3}
		\includegraphics[width=0.29\linewidth]{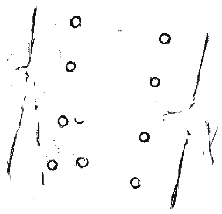}
	}
	\subfigure[UWB4]{
		\label{figure:UWB4}
		\includegraphics[width=0.29\linewidth]{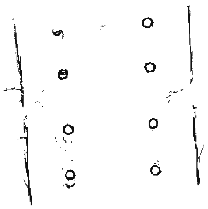}
	}
	\subfigure[combination of UWB3 and UWB4]{
		\label{figure:UWB34}
		\includegraphics[width=0.29\linewidth]{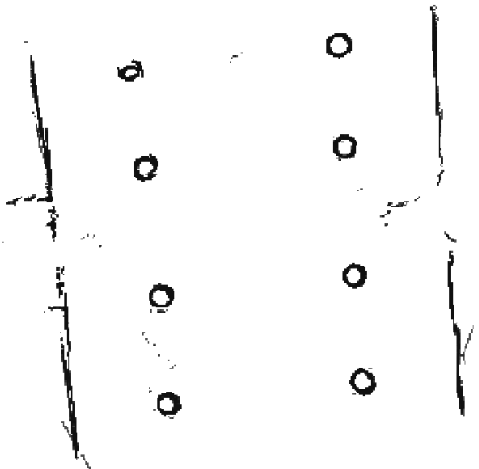}
	}	
	
	\subfigure[combination of UWB2, UWB3 and UWB4]{
		\label{figure:UWB234}
		\includegraphics[width=0.29\linewidth]{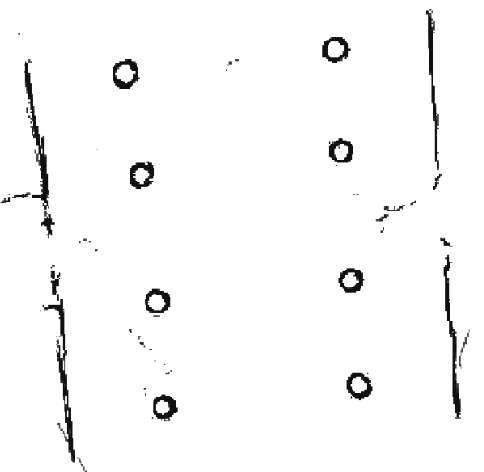}
	}	
	\subfigure[four UWB]{
		\label{figure:UWB5}
		\includegraphics[width=0.29\linewidth]{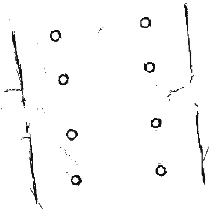}
	}
	\caption[mapping results]{Impact of the number of UWB on the result accuracy.
	}

	\label{figure:Impact of the number of UWB}
\end{figure} 

\renewcommand{\arraystretch}{1.5} 
\begin{table}[H] \huge
	\centering  
	\caption{Influence of different UWB nodes on the mean error.}
	\label{tab:MSE of different UWB anchor}  
	
	\scalebox{0.9}{
		\scalebox{0.369}{
			\begin{tabular}{|c|c|c|}

				\hline
				\textbf{Different UWB nodes} & \textbf{\begin{tabular}[c]{@{}c@{}}The number of\\ UWB edges \end{tabular}} & \textbf{Mean mapping error(m)} \\ \hline
				{\Huge UWB1-only}                  & {\Huge731}                                    & {\Huge0.210}                  \\ 
				{\Huge UWB2-only}                 & {\Huge733}                                    & {\Huge0.205}                  \\ 
				{\Huge UWB3-only}                  & {\Huge755}                                    & {\Huge0.551}                  \\ 
				{\Huge UWB4-only}                  & {\Huge612}                                    & {\Huge0.345}                  \\
				{\Huge UWB3 and UWB4}              & {\Huge1367}                                   & {\Huge0.275}                  \\ 
				{\Huge UWB2, UWB3, and UWB4}       & {\Huge2100}                                   & {\Huge0.202}                  \\ 
				{\Huge Four UWB}                   & {\Huge2831}                                   & {\Huge0.199}                  \\ \hline
			\end{tabular}
	}}
\end{table}
\subsection{Impact of Different UWB nodes on Mapping Accuracy}
In this section, the influence of the number of UWB nodes on the mapping accuracy is investigated. The results are shown in Fig.~\ref{figure:Impact of the number of UWB} and Table~\ref{tab:MSE of different UWB anchor}. We choose moving distance threshold $\epsilon$ =9.0, matching score threshold $\sigma$=0.1, and distance threshold $d$=0.1. We set the number of iterations to 300. It should be indicated that different UWB nodes have different impacts on the constructed map. When only one UWB node is utilized, the UWB edges are 731, 733, 755 and 612. However, when four UWB nodes are used, the edge of the graph is 2831. Fig.~\ref{figure:UWB5} shows the best mapping result when four UWB nodes are used to construct the map. Furthermore, Fig.~\ref{figure:UWB1} and Fig.~\ref{figure:UWB2} present the maps created by UWB1 and UWB2, respectively. It is observed that they are quite similar to the one created by four UWBs (see Fig.~\ref{figure:UWB5}). However, the maps obtained from UWB3 and UWB4 (i.e., Fig.~\ref{figure:UWB3} and Fig.~\ref{figure:UWB4}) are quite different from the real environment.

Due to dynamic obstacles such as pedestrians and non-line of sight (NLOS) signals from the environment, the mean mapping error of UWB3 and UWB4 is larger than mapping error of UWB1 and UWB2. Table~\ref{tab:MSE of different UWB anchor} also shows that mapping accuracy improves with more UWB. For example, the mapping error with a combination of UWB3 and UWB4 is 0.275m, which is better than UWB3 (0.551m) and UWB4 (0.345m) alone. Nonetheless, we can observe that the mapping error with the fusion of four UWBs (0.199m) is smaller than the fusion of two UWBs (0.275m using UWB3 and UWB4) or three UWBs (0.202m using UWB2, UWB3, and UWB4).  Due to the block of the obstacles, we might not always observe the four UWB nodes at the same time. In addition, as shown in Table~\ref{tab:MSE of different UWB anchor}, we obtain a good mapping accuracy with two UWB nodes. For different scenarios, we need to make sure that the mapping space is covered by at least two UWB nodes. More UWB nodes will give a better mapping accuracy (for example, a mapping accuracy of 0.202m with the fusion of three UWB nodes, i.e., UWB2, UWB3, and UWB4). But optimization with more UWBs requires more computational time.  We suggest to uniformally distribute the UWB nodes in an environment to ensure the maximum coverage of the environment.  


\renewcommand{\arraystretch}{1.5} 
\begin{table}[H] \Huge
	\caption{Impact of different $\epsilon$ on the result accuracy.}
	\label{tab:Impact of different d}
	\centering
	\scalebox{0.56}{
		\scalebox{0.52}	{
			\begin{tabular}{|c|c|c|c|c|}
				\hline
				\textbf{\bm{$\epsilon(m)$}} & \textbf{\begin{tabular}[c]{@{}c@{}}Time(in seconds) required for \\ LiDAR loop closures\end{tabular}} & \textbf{\begin{tabular}[c]{@{}c@{}}The number of\\  LiDAR loop\end{tabular}} & \textbf{Mean mapping error(m)} \\ \hline
				{\Huge0.0}                    & {\Huge12436}                                                                                                                          & {\Huge3311}                                                                         & {\Huge0.501}                  \\
				{\Huge2.0}                    & {\Huge65}                                                                                                 & {\Huge111}                                                                           & {\Huge0.493}                  \\ 
				{\Huge4.0}                      &                  {\Huge180}                                           & {\Huge98}                                                                                    & {\Huge0.484}                  \\ 
				{\Huge6.0}                    & {\Huge321}                                                                                                   & {\Huge112}                                                                       & {\Huge0.280}                  \\ 
				{\Huge8.0}                    & {\Huge495}                                                                                                   & {\Huge252}                                                                      & {\Huge0.210}                  \\ 
				{\Huge9.0}                   & {\Huge523}                                                                                                  & {\Huge127}                                                                     & {\Huge0.199}                 \\ 
				{\Huge10.0}                    & {\Huge563}                                                                                                   & {\Huge118}                                                                       & {\Huge0.223}                  \\ 
				{\Huge11.0}                    & {\Huge594}                                                                                                  & {\Huge102}                                                                      & {\Huge0.258}                  \\ 
				{\Huge12.0}                    & {\Huge600}                                                                                                   & {\Huge100}                                                                       & {\Huge0.292}                  \\ \hline
			\end{tabular}
	}}
\end{table}

\subsection{Impact of Distance Threshold $\epsilon$ on Mapping Accuracy}
In this section, the number of LiDAR loop closures, the required time for detecting the LiDAR loop closure, and the mapping accuracy for different thresholds $\epsilon$ are investigated. Table~\ref{tab:Impact of different d} presents the obtained results. We choose matching score threshold $\sigma$=0.1, distance threshold $d$=0.1, and the number of iterations as 300. All four UWB nodes are used.  It is observed that when $\epsilon$ is set to zero, the proposed approach is equivalent to the conventional ICP. Meanwhile, it is found that the conventional ICP algorithm cannot detect the correct LiDAR loop closure in a feature-less environment. Accordingly, a large number of false LiDAR loop closures are created, which remarkably reduces the mapping accuracy. Therefore, it is important to find the optimal value of $\epsilon$ to reduce the LiDAR loop detection time and improve the quality and accuracy of the map. Meanwhile, little environmental information is involved in the local map if the robot moves a short distance, resulting in many incorrect loop closures. On the other hand, we will ignore many correct LiDAR loop closure if the traveled distance is too large. Table~\ref{tab:Impact of different d} indicates that the best result is obtained when $\epsilon$ is set to 9.0.

\subsection{Impact of Number of Iterations on Mapping Accuracy}
In this section, it is intended to investigate the influence of different number of iterations on the mapping accuracy and required process time of LiDAR loop closures. We choose moving distance threshold $\epsilon$ =9.0, matching score threshold $\sigma$=0.1, and distance threshold $d$=0.1. All four UWB nodes are used for the optimization. It is clear that as the number of iterations in the ICP algorithm increase, the required processing time for the LiDAR loop closures detection increases. On the other hand, an appropriate LiDAR loop closure cannot be detected for too small number of iterations, which results in poor map quality. Therefore, it is of significant importance to perform an engineering trade-off and find an appropriate number of iterations to reduce the computational expense and processing time for the LiDAR loop closure and improve the mapping accuracy and quality.

\renewcommand{\arraystretch}{1.5} 
\begin{table}[H] \huge
	
	\centering  
	
	\caption{Impact of the number of iterations on the mapping accuracy.}
	\label{tab:Impact of the number of iteration}  
	
	\scalebox{0.325}{
		{
			\begin{tabular}{|c|c|c|c|}
				\hline
				\textbf{Iterations} & \textbf{\begin{tabular}[c]{@{}c@{}}Time(in seconds) required \\ for LiDAR loop closures\end{tabular}} & \textbf{\begin{tabular}[c]{@{}c@{}}The number of \\  LiDAR loop closures\end{tabular}} & \textbf{Mean mapping error(m)} \\ \hline
				{\Huge10}                  & {\Huge80}                                                                                                    & {\Huge90}                                         & {\Huge0.287}                  \\
				{\Huge50}                  & {\Huge327}                                                                                                   & {\Huge110}                                        & {\Huge0.245}                  \\ 
				{\Huge80}                  & {\Huge433}                                                                                                   & {\Huge116}                                        & {\Huge0.240}                  \\ 
				{\Huge100}                 & {\Huge440}                                                                                                   & {\Huge120}                                        & {\Huge0.235}                  \\
				{\Huge300}                 & {\Huge523}                                                                                                   & {\Huge127}                                        & {\Huge0.199}                  \\ 
				{\Huge500}                 & {\Huge625}                                                                                                   & {\Huge127}                                        & {\Huge0.199}                  \\ 
				{\Huge1000}                & {\Huge877}                                                                                                   & {\Huge127}                                        & {\Huge0.199}                  \\ \hline
			\end{tabular}
	}}
\end{table}

Table~\ref{tab:Impact of the number of iteration} indicates that for 10 iterations, the required time to process the LiDAR loop closure is about 80 seconds and the number of the LiDAR loop closure reaches 90. It is observed that as the number of iterations increases, the time and the number of the corrected LiDAR loop closure gradually increase, while the mean error gradually decreases. It indicates that for 300 iterations, the number of the LiDAR loop closure and mean error does not change, while the detection time increases. It is inferred that the performed numerical simulation converges with 300 iterations. Therefore, 300 iterations are considered in the calculations.
\begin{figure}[H]
	\centering
	\subfigure[Mean error for different $\sigma$ and $d$ values]{
		\label{figure:1}
		\includegraphics[width=0.7\linewidth]{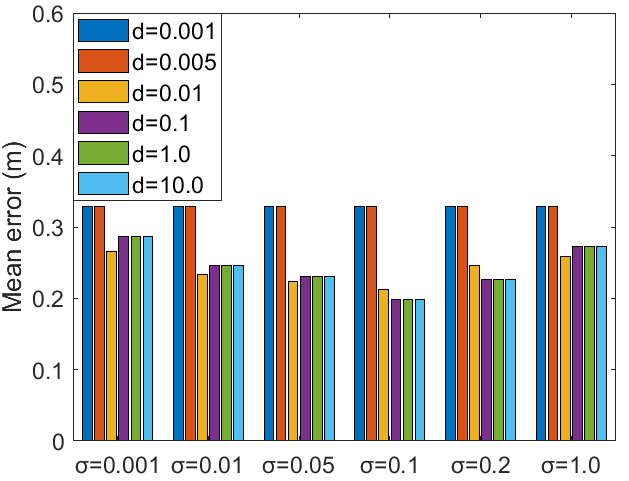}
	}
	\subfigure[Number of LiDAR constraints for various $\sigma$ and $d$]{
		\label{figure:2}
		\includegraphics[width=0.7\linewidth]{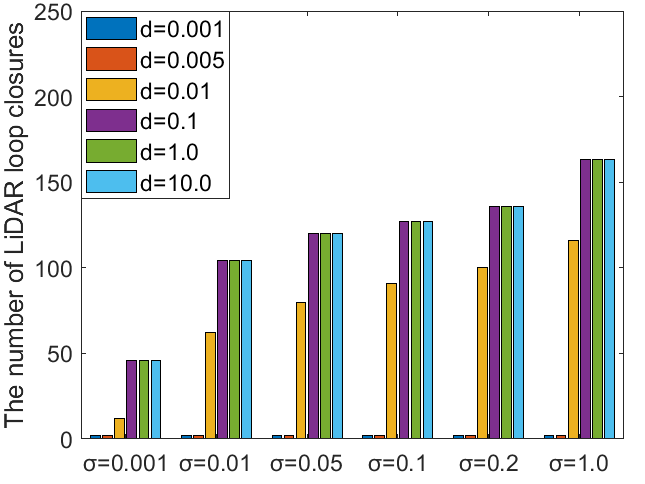}
	}
	
	\caption[mapping results]{Impact of $\sigma$ and $d$ on the mapping accuracy.
	}
	\label{figure:Impact of The Matching Score($s$) and Distance Threshold(d)}
\end{figure} 
\subsection{Impact of the Matching Score Threshold $\sigma$ and Distance Threshold $d$ on the Mapping Accuracy}
In this section, the number of LiDAR loop closure and mapping accuracy with different distance thresholds $d$ and matching score threshold $\sigma$ is studied. Fig.~\ref{figure:Impact of The Matching Score($s$) and Distance Threshold(d)} shows the obtained results in this regard. We set moving distance threshold $\epsilon$ =9.0 and the number of iterations as 300. We use all UWB nodes for the optimization.
Fig.~\ref{figure:Impact of The Matching Score($s$) and Distance Threshold(d)} shows that too large or too small values of $d$ and $\sigma$ reduce the mapping accuracy. It is found that for low threshold of $\sigma$ and $d$, several correct LiDAR loop closures are ignored. On the other hand, for too large $\sigma$ and $d$ values, several wrong closures appear. Small values of $\sigma$ and $d$ are not capable of identifying potential loops so that a small number of LiDAR-based loops are obtained. The number of LiDAR loop closure does not increase only if the $d$ increases to 0.1m. Based on the performed investigation, it is concluded that the best mapping quality can be obtained when both parameters $\sigma$ and $d$ are set to 0.1 and the number of the LiDAR loop closure is 127.

\subsection{Computational Time}
Finally, the time consumption at each stage of the proposed approach is evaluated. Table~\ref {tab:Evaluation of the computational time (in seconds) in each stage} presents the obtained results at different stages with different UWB nodes. It should be indicated that in the present study, an Intel Core i3-2328M CPU with 2.20 GHz frequency and 4 GB RAM is employed to process the measurements. Table~\ref {tab:Evaluation of the computational time (in seconds) in each stage} shows that the entire data processing with four UWB nodes takes 528.947s (2.657 + 547.321 + 2.819), which is almost 4 times faster than the data recording stage. Moreover, the optimization of the graph takes less than 3s. Optimization times for the first and second optimizations are 2.657s and 2.819s when four UWB nodes are used. Compared with using four UWB nodes, the processing time is reduced when one UWB node is used. In the current offline implementation, the similarity of two LiDAR measurements is computed to find the potential loop closures.  Table~\ref{tab:Evaluation of the computational time (in seconds) in each stage} indicates that the LiDAR loop closure module consumes too much time. Moreover, it takes a longer time to run the algorithm, which originates from the increasing number of vertices and edges in the graph.

\renewcommand{\arraystretch}{1.6} 

\begin{table}[H] \huge
	\centering  
	
	\caption{Evaluation of the computational time (in seconds) in each stage with different numbers of UWB nodes used}
	\label{tab:Evaluation of the computational time (in seconds) in each stage}  
	\scalebox{0.7}{
		\scalebox{0.6}{
			\begin{tabular}{|c|c|c|}
				\hline
				\multirow{2}{*}{\textbf{Stage}}                    & \multicolumn{2}{c|}{\textbf{Duraion(s)}}                             \\ \cline{2-3} 
				& \multicolumn{1}{l|}{Four UWB nodes} & \multicolumn{1}{l|}{UWB1-only} \\ \hline
				Data recording                                      & 1909.648                            & 1909.648                       \\ 
				First optimization                                 & 2.657                               & 1.450                          \\ 
				\multicolumn{1}{|l|}{LiDAR loop closure detection} & 523.471                             & 413.670                        \\ 
				Second optimization                                & 2.819                               & 2.342                          \\ \hline
			\end{tabular}
	}}
\end{table}

\subsection{Mapping Results for Different Scenarios}
\begin{figure}[H]
	\centering
		\subfigure[A snapshot of the experiment in a corridor environment]{
	\label{figure:complex real environment}
		\includegraphics[width=9cm]{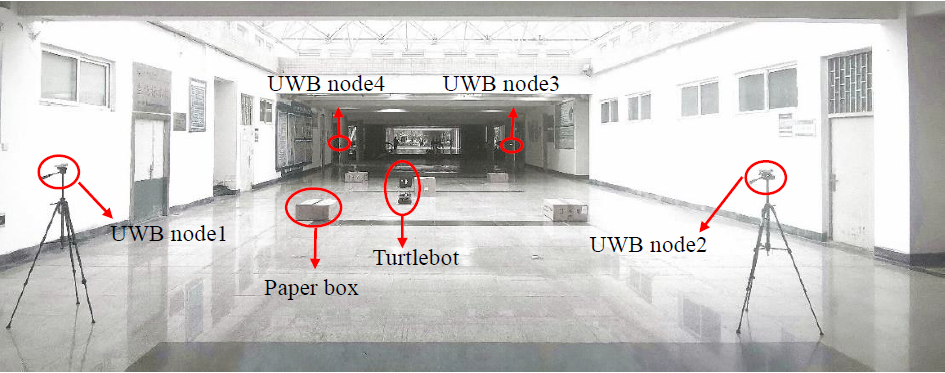}
	}
	\subfigure[Map created by GMapping]{
		\label{figure:map created by Gmapping}
		\includegraphics[width=0.3\linewidth]{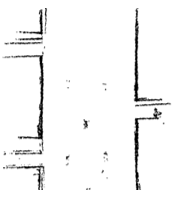}
	}
	\subfigure[Map created by our approach]{
		\label{figure:Map created by Map created by the proposed approach}
		\includegraphics[width=0.27\linewidth]{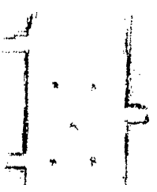}
	}
	\caption[Obtained maps from different algorithms.]{Maps created from different approaches in a corridor environment.
	
	}
	\label{figure: no feature result}
\end{figure}

\begin{figure}[H]
	\centering
	\subfigure[A snapshot of the complex indoor environment]{
		\label{figure:complex environmentg}
		\includegraphics[width=5.5cm]{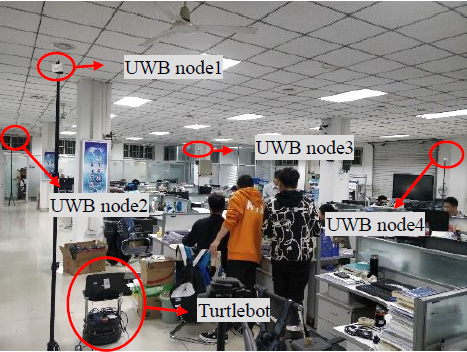}
	}
	\subfigure[Map created by GMapping]{
		\label{figure:Map created by Gmapping}
		\includegraphics[width=0.44\linewidth]{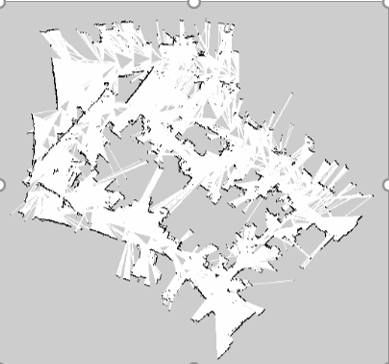}
	}
	\subfigure[Map created by our approach]{
		\label{figure: map created by the proposed approach}
		\includegraphics[width=0.48\linewidth]{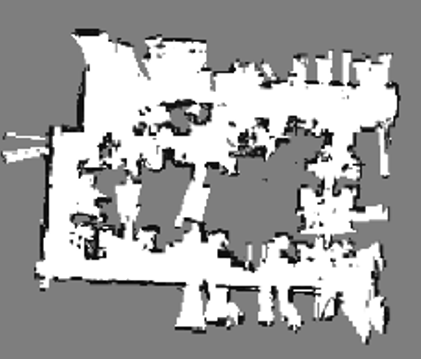}
	}
	\caption[Maps from different algorithms.]{Maps created from different approaches in a complex indoor environment.
	}
	\label{figure:Complex result}
\end{figure}

We additionally performed two different experiments to verify the validity of our proposed approach: one experiment in a corridor with less features (with a size of $12m\times26m$) as shown in Fig.~\ref{figure: no feature result}, and another experiment in an indoor complex environment (our laboratory with a size of $15m\times20m$) as shown in Fig.~\ref{figure:Complex result}. Four UWB nodes are placed in the test environment, and the locations of UWB nodes are not known.

Fig.~\ref{figure: no feature result} and Fig.~\ref{figure:Complex result} give a comparison of the maps created by different approaches, namely GMapping and our approach. The core of SLAM is to use the sensors equipped on the robot (i.e., LiDAR) to correct the accumulated odometry error. As it can be seen from these two experiments, GMapping produces poor maps in these environments, as the short-range LiDAR fails to correct the large odometry error due to the lack of good features, which leads to substantial offsets of the corridors and walls as shown in Fig.~\ref{figure:map created by Gmapping} and Fig.~\ref{figure:Map created by Gmapping}. In contrast, the maps produced by our approach (see Fig.~\ref{figure:Map created by Map created by the proposed approach} and Fig.~\ref{figure: map created by the proposed approach}) are better than GMapping. In our approach, UWB can be considered as additional attributes to LiDAR, which allows to correct the large odometry error in feature-less and complex indoor environments, where the conventional GMapping fails to find enough features for localization. 	With an increasing of LiDAR ranging, the LiDAR can cover more areas and provide more features of the environment, as a result the traditional algorithms such as GMapping is able to correct the accumulated error of odometry and achieves a good mapping accuracy without the fusion of UWB.

\section{Conclusion}
\label{Conclusion}
In the present study, a fusion strategy is proposed using measurements of UWB, odometry, and low-cost 2D LiDAR to construct an accurate map in a feature-less environment. UWB and odometry information is fused into a graph optimization to remove the cumulative error of the odometry, and obtain the initial trajectory. Then, the LiDAR loop closure information is integrated by constructing local sub-maps to further constrain the robot trajectory. Obtained results demonstrate that the proposed method can effectively resolve the limitations of the conventional ICP algorithm in the feature-less environment, and improve the accuracy of the trajectory estimation and the mapping quality. Experimental results show that the accuracy and quality of the map constructed by the proposed algorithm are higher than that of the conventional GMapping algorithm using odometry and LiDAR. More specifically, mapping error of the proposed approach is 85.5\% less than that of the conventional GMapping algorithm. In the near future, it is intended to enhance the accuracy of the map in large-scale environments to satisfy the requirements of real industrial applications. Moreover, indoor navigation and path planning will be evaluated by applying the map constructed by the proposed system.

\begin{IEEEbiography}[{\includegraphics[width=1in,height=1.25in,clip,keepaspectratio]{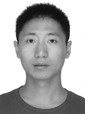}}]{Ran Liu} received the B.S. degree from the Southwest University of Science and Technology,
	Mianyang, China, in 2007, and the Ph.D. degree from the University of Tüebingen, Tübingen, Germany, in 2014, under the supervision of Prof. Dr. A. Zell and Prof. Dr. A. Schilling. Since 2014, he has been a Research Fellow under
	the supervision of Prof. C. Yuen with the MIT International Design Center, Singapore University of Technology and Design, Singapore. His research interests include robotics, indoor positioning, and SLAM.
\end{IEEEbiography}

\begin{IEEEbiography}[{\includegraphics[width=1in,height=1.25in,clip,keepaspectratio]{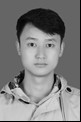}}]{Yongping He}
	received the B.S. degree from Southwest University of Science and Technology, Mianyang, China, in 2018, and is currently studying for the MA.Sc degree at Southwest University of Science and Technology. His research interests include SLAM, UWB positioning.
\end{IEEEbiography}

\begin{IEEEbiography}[{\includegraphics[width=1in,height=1.25in,clip,keepaspectratio]{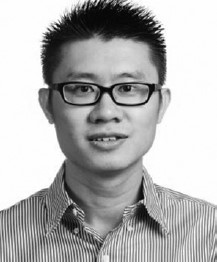}}]{Chau Yuen} received the B.Eng. and Ph.D. degrees from Nanyang Technological University, Singapore, in 2000 and 2004, respectively. He was a Postdoctoral Fellow with Lucent Technologies Bell Labs, Murray Hill, NJ, USA, in 2005. He was a Visiting Assistant Professor with the Hong Kong Polytechnic University, Hong Kong, in 2008. From 2006 to 2010, he was a Senior Research Engineer with the Institute for Infocomm Research, Singapore, where he was involved in an industrial project on developing an 802.11n Wireless LAN System, and participated actively in 3GPP Long Term Evolution (LTE) and LTE-Advanced Standardization. He has been with the Singapore University of Technology and Design, Singapore, as an Assistant Professor since 2010. Dr. Yuen is a recipient of the Lee Kuan Yew Gold Medal, the Institution of Electrical Engineers Book Prize, the Institute of Engineering of Singapore Gold Medal, the Merck Sharp and Dohme Gold Medal, and twice the recipient of the Hewlett Packard Prize, and the IEEE Asia–Pacific Outstanding Young Researcher Award in 2012. He serves as an Editor for the IEEE TRANSACTION ON COMMUNICATIONS and the IEEE TRANSACTIONS ON VEHICULAR TECHNOLOGY, and was awarded the Top Associate Editor from 2009 to 2015.
\end{IEEEbiography}

\begin{IEEEbiography}[{\includegraphics[width=1in,height=1.25in,clip,keepaspectratio]{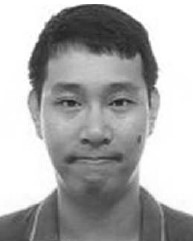}}]{Billy Pik Lik Lau } received the B.Sc. degree and the M.Phil. degree in computer science, with a focus on improving cooperation rate between agents in multiagents systems, from Curtin University, Perth, WA, Australia, in 2010 and 2014, respectively. He is currently pursuing the Ph.D. degree with the Singapore University of Technology and Design, Singapore, under Dr. Yuen Chau’s supervision. His current research interests include smart city, Internet of Things, big data analysis, data discovery, and unsupervised machine learning.
\end{IEEEbiography}

\begin{IEEEbiography}[{\includegraphics[width=1in,height=1.25in,clip,keepaspectratio]{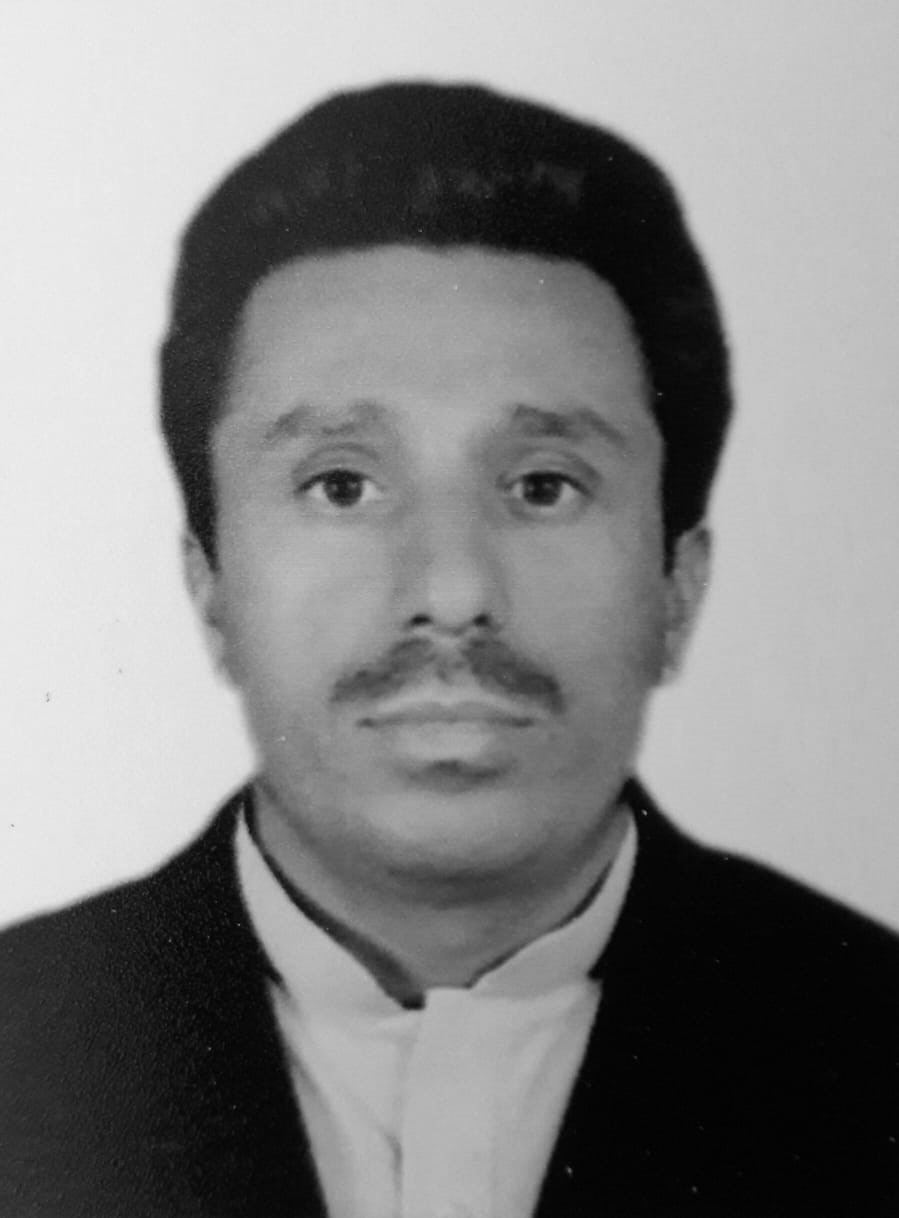}}]{Rashid Ali}is currently pursuing the Ph.D.degree with the School of Information Engineering, Southwest University of Science and Technology, Mianyang, China	His research interests include Multi-sensor vehicle/robot positioning and localization.
\end{IEEEbiography}

\begin{IEEEbiography}[{\includegraphics[width=1in,height=1.25in,clip,keepaspectratio]{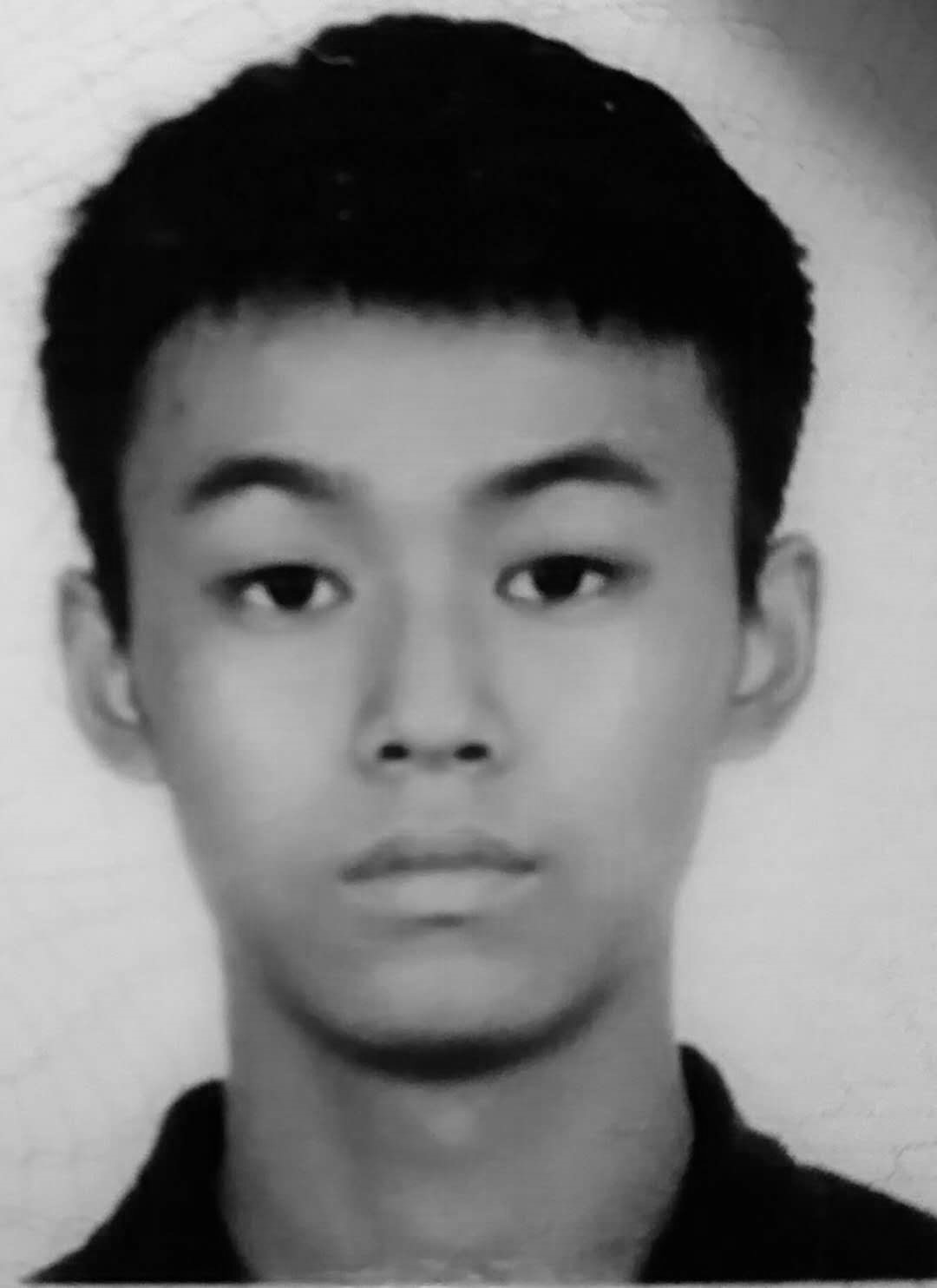}}]{Wenpeng Fu} received the B.S. degree from Southwest University of Science and Technology, Mianyang, China in 2017, and is currently studying for the MA.Sc degree at Southwest University of Science and Technology. His research interests include SLAM, UHF RFID localization.
\end{IEEEbiography}

\begin{IEEEbiography}[{\includegraphics[width=1in,height=1.25in,clip,keepaspectratio]{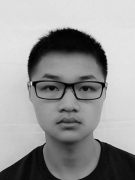}}]{Zhiqiang Cao } received the B.S. degree from the Southwest University of Science and Technology, Mianyang, China, in 2019. He is currently pursuing the MA.Sc degree with the school of Information Engineering, Southwest University of Science and Technology. His research interests include indoor localization, SLAM, UWB.
\end{IEEEbiography}






\end{document}